\documentclass[letterpaper]{article} % DO NOT CHANGE THIS
\usepackage[]{aaai23}
\usepackage{times}  % DO NOT CHANGE THIS
\usepackage{helvet}  % DO NOT CHANGE THIS
\usepackage{courier}  % DO NOT CHANGE THIS
\usepackage[hyphens]{url}  % DO NOT CHANGE THIS
\usepackage{graphicx} % DO NOT CHANGE THIS
\usepackage{comment}
\usepackage{natbib}  % DO NOT CHANGE THIS AND DO NOT ADD ANY OPTIONS TO IT
\usepackage{caption} % DO NOT CHANGE THIS AND DO NOT ADD ANY OPTIONS TO IT
\usepackage{algorithm}
\usepackage{algorithmic}
\usepackage{amsmath}
\usepackage{amssymb}
\usepackage{bm}
\usepackage{upgreek}
\usepackage{subcaption}
\usepackage{xcolor}
\usepackage{breakcites}

\usepackage{newfloat}
\usepackage{listings}
\DeclareCaptionStyle{ruled}{labelfont=normalfont,labelsep=colon,strut=off} % DO NOT CHANGE THIS
\floatstyle{ruled}
\newfloat{listing}{tb}{lst}{}
\floatname{listing}{Listing}
\newcommand{\hidecomment}[1]{}

\title{\textsc{LogicRank}: \\ Logic Induced Reranking for Generative Text-to-Image Systems}

\author{Björn Deiseroth$^{*\filledstar\smallstar}$, Patrick Schramowski$^{*\filledstar}$, Hikaru Shindo$^*$,  \\ Devendra Singh Dhami$^{*\filledstar}$, Kristian Kersting$^{*\filledstar}$}

\affiliations{
$^*$Technische Universität Darmstadt \hspace{.6cm} 
$^\filledstar$Hessian Center for AI (hessian.AI) \hspace{.6cm} 
$^\smallstar$Aleph Alpha
}

\usepackage{MnSymbol}
\newcommand*{\Scale}[2][4]{\scalebox{#1}{$#2$}}%

\begin{document}
\maketitle

\begin{abstract}
Text-to-image models have recently achieved remarkable success with seemingly accurate samples in photo-realistic quality. 
However as state-of-the-art language models still struggle evaluating precise statements consistently, so do language model based image generation processes.  
In this work we showcase problems of state-of-the-art text-to-image models like DALL-E with generating accurate samples from statements related to the \textit{draw bench} benchmark. Furthermore we show that CLIP is not able to rerank those generated samples consistently.
To this end we propose LogicRank, a neuro-symbolic reasoning framework that can result in a more accurate ranking-system for such precision-demanding settings. LogicRank integrates smoothly into the generation process of text-to-image models and moreover can be used to further fine-tune towards a more logical precise model.
\end{abstract}

%\begin{keywords}
%\end{keywords}

\section{Introduction}
\begin{figure}[t]
\includegraphics[width=\linewidth,trim={0 8cm 20.2cm 0},clip]{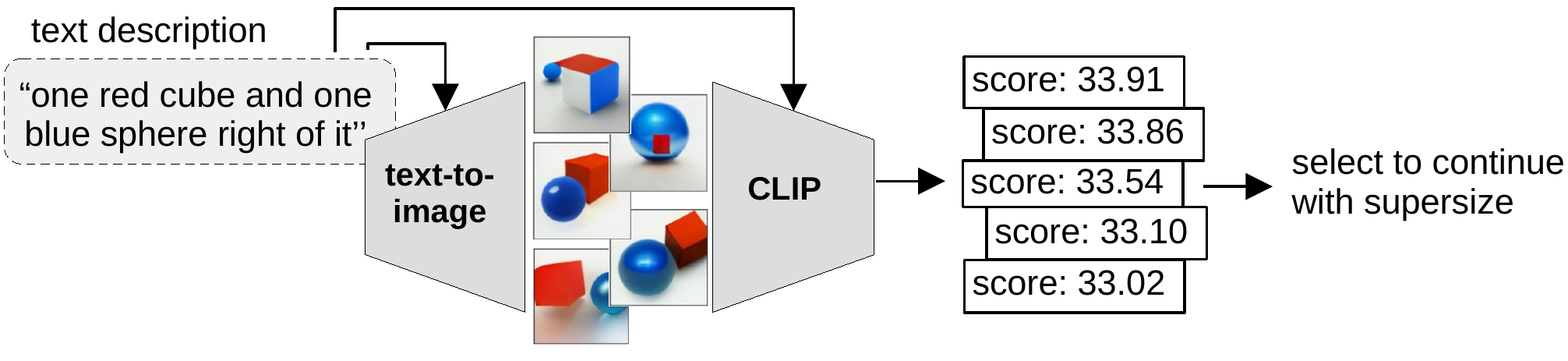}
\caption{First step of the DALL-E pipeline: A text description of an image is inserted to generate samples, which get reranked and selected by CLIP, before selecting candidates for the final image-polishing steps.}
\label{fig:dalle_workflow}
\end{figure}

%\section{A bottleneck of\\ generative language models}
\label{sec:intro}
Transformer based Language models (LMs), such as BERT~\cite{devlin2018bert} and GPT~\cite{gpt}, have achieved success in several natural language understanding (NLU) tasks. However, there are growing concerns that these LMs cannot perform true logical reasoning~\cite{lm_comprehension, neural_proof}. 
Naturally these problems will be inherited in multi-modal setups, as in the case of text-to-image generation tasks with models like DALL-E~\cite{dalle,dalle2} or imagen~\cite{imagen}. The authors of DALL-E already mentioned this impairment in the context of geometric scene descriptions.
Some approaches, such as DALL-E,  suggest to sample several images within the forward pass and use text-image embedding models like CLIP to rerank generated images for best matching candidates to proceed with. This workflow is depicted in Fig.~\ref{fig:dalle_workflow}. After selection, the best ranked image may be further polished e.g. with supersizing or sharpening procedures. That post-processing is not discussed further in this work. 

However using CLIP as a reranker for DALL-E may not yield the expected performance for the following two reasons:
1) images generated with machine learning models can produce significant artifacts on other machine learning models, in particular as the generated data is out of training distribution
2) the generating and reranking models are both trained on a similar dataset and loss, therefore assumably containing similar bottlenecks, such as sufficient representations of details. 

In Fig.~\ref{fig:dalle_workflow} we show the top-5 candidates that were reranked by CLIP from 200 generated samples for the prompt ``one red cube and one blue sphere right of it'' with DALL-E mini\footnote{We used the huggingface version ``dalle-mini/dalle-mega''~\cite{dalle-mega}.}.  It can be observed that concepts like matching certain colors to the right shape, as well as the object orientation are often ignored. 
We thus call for a different validation approach as a reranker for the sampling process. 

To this end, we propose \emph{Logic induced Reranking for generative Text-to-Image Systems} (LogicRank) which is a Neuro-Symbolic Forward Reasoning approach to reason and evaluate generated samples for precise statements.
LogicRank is able to consistently find the \textit{needle in a haystack}, the logical correct generation under hundreds of examples.
With such ``reasoned feedback'', we hope to enrich the generation process of such models, to obtain more precise results for logical complex statements. Moreover it could be applied to further fine-tune the generating model.

Overall, we make the following contributions: 
 1) Introduction of LogicRank, a logical ranking system integratable into  text-to-image generation systems.
 2) Empirical Comparison of LogicRank and CLIP as ranking systems for image selection tasks.

\begin{figure*}[t]
\includegraphics[width=\textwidth,trim={0 2.8cm 2.5cm 0},clip]{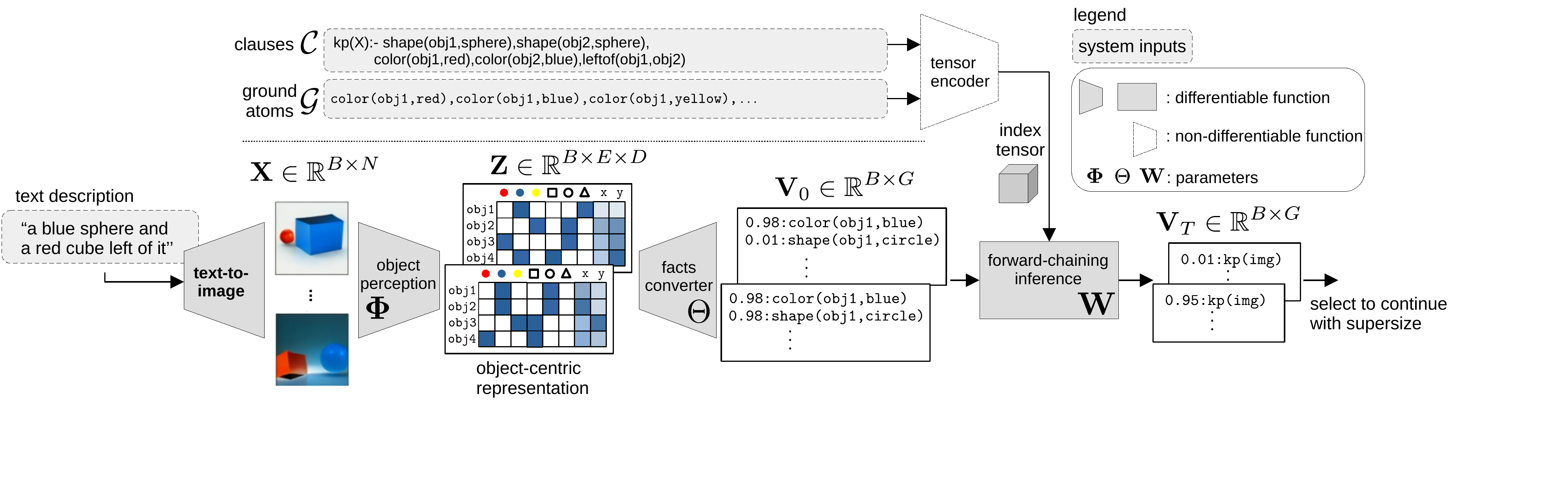}
\captionof{figure}{Proposed workflow of LogicRank. A text-to-image model generates images given text in natural language. The generated images are fed into  a visual-perception model to factorize the input raw images in terms of objects. The factorized output is converted into probabilistic facts, where each fact is associated with a probabilistic value. Finally, forward-chaining inference is performed to compute the logical entailment softly from the probabilistic facts and weighted rules. The final prediction is computed based on the entailed facts.}
\label{fig:workflow}
\end{figure*}

\section{LogicRank as a Reasoned Reranker}
\label{sec:reranker}
As a reranker for logical statements, we propose LogicRank. 
It is an abstract framework that consumes a generative text-to-image model, object detectors, text prompts and their translation into logical rules, and outputs generated samples with their prompt-matching probabilities.  
The workflow is  depicted in Fig.~\ref{fig:workflow}. 
In the following paragraphs we first give a motivating example before  describing the separate components in more detail.

\paragraph*{Example.}

In this study we use the DALL-E mini  text-to-image model to evaluate the framework. The DALL-E approach~\cite{dalle} itself generates several samples, which are subsequently reranked. 
The iterative sample and rerank process is essential as it maximizes the evidence lower bound on the joint likelihood of the model distribution over images, captions and tokens.
Fig.~\ref{fig:dalle_workflow} depicts the original reranking process utilizing the pre-trained CLIP model as a reranker. CLIP embeds the generated images as well as the given text prompt and computes the similarity of both modalities. The similarity score is used for the final reranking.
We observe that using CLIP\footnote{We used the huggingface version ``openai/clip-vit-large-patch14''.} as a reranker fails for many types of queries, e.g. positional precise statements. 
To encounter this drawback, we propose the workflow as follows. 
%The generated images of DALL-E, and the input text translated into a logical rule are fed into NSFR. It calls DETR as $f_\text{per}^{\bm{\Phi}}$ to detect relevant atoms, and as previously described to derive the normalized probability estimate $\Tilde{p}(\mathbf{y}|\mathbf{X})$.

The text statement and its translation into a logical rule is passed as input as depicted in Fig.~\ref{fig:workflow}. 
We first let the text-to-image model generate a set of corresponding  samples to the text prompt. 
An object detection model generates an object centric representation of the images.
We found that little fine-tuning of DETR~\cite{detr} works well for attribute detection in generated samples.
Finally this object representation along with the logical statement rule is passed into the Neuro-Symbolic Forward Reasoner (NFSR)~\cite{nsfr} as described below, to derive  the normalized probability estimate.

\begin{figure}[t]
\vspace{-20pt}
\begin{table}[H]
 \begin{tabular}{p{3.5cm}p{3.3cm}}
\raisebox{-\totalheight}{\includegraphics[width=\linewidth]{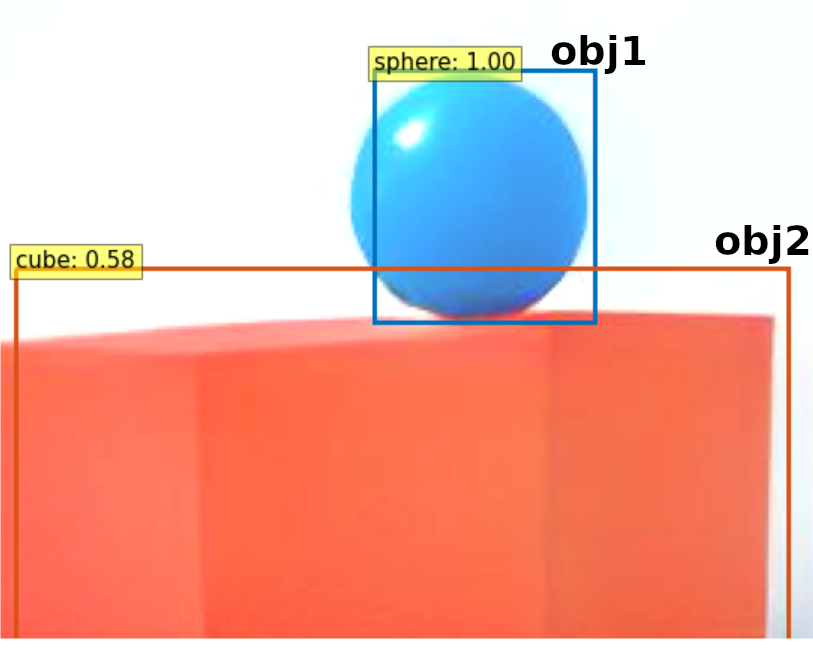}} & 
\hspace{-50pt}
\[\Scale[.95]{\begin{array}{l}
{shape(obj1, \textrm{sphere}): 1,}\\ 
{color(obj1, \textrm{blue}): 0.95:,}\\
{shape(obj2, \textrm{cube}): 0.58,}\\
{color(obj2, \textrm{red}): 0.83,}\\
{position(obj1, obj2, \textrm{above}): 1}\\
\Rightarrow {kp: 0.86}
\end{array}}\]
 \end{tabular}
\end{table}
\vspace{-15pt}
\caption{DETR object detection of a DALL-E mini-generated image, after being finetuned on CLEVR as described in text. Next to it, the NSFR discovered rule for ``a blue sphere on top of a red cube'' ($kp$) and the derived probability estimates.
\label{fig:finetuned}}
\end{figure}

Fig.~\ref{fig:finetuned} shows an inference example. To evaluate the prompt ``a blue sphere on top of a red cube'', we use the  logical form: $kp\ \textrm{:-}\ 
shape(obj1, sphere),\ \allowbreak color(obj1, blue),\ \allowbreak shape(obj2, cube),\ \allowbreak color(obj2, red),\ \allowbreak position(obj1, obj2,\allowbreak above)$. The image shows an example of detected shape atoms with their probabilities thru DETR. For each bounding-box we further induce the color attributes and a positional relation. 
All derived attribute estimates are shown next to the figure. 
As a final result from the reasoner, we get the  conclusion $\Tilde{p}(kp|\mathbf{X}) = 0.86= \sqrt[5]{0.46}$ for the rule in question.

This example already indicates a benefit of NSFR compared to  CLIP-based reranking. 
We are not only confronted with the final result but also get detailed reasoned feedback, e.g about dubious attributes to improve on.

\paragraph*{DETR.}
As an object detector we use DETR~\cite{detr}\footnote{We used the huggingface version ``facebook/detr-resnet-101''~\cite{detr}}. 
For the problem space of evaluating the performance of text-to-image models on placing objects in scenes properly, we further finetune DETR as follows: 
We use the CLEVR framework~\cite{clevr} to generate 1000 images in a ``DALL-E mini style'', with adjusted colors and proportions. CLEVR offers a dynamic framework for  generating synthetic images and its text descriptions for scenes with randomly placed objects.
We further modify Contrast and Brightness levels randomly, to obtain a more robust classifier. DETR is finetuned for 3000 steps on this dataset.

%\noindent

\paragraph*{Neuro-Symbolic Forward Reasoner.}
The NSFR is an end-to-end differentibale reasoning framework.
%The key idea is to combine differentiable forward-chain reasoning with object-centric (deep) learning. Differentiable forward-chain reasoning computes logical entailments smoothly, i.e., it deduces new facts from given facts and rules in a differentiable manner.
%The object-centric learning approach factorizes raw inputs into representations in terms of objects. %Thus, it allows us to provide a consistent framework to perform the forward-chaining inference from raw inputs.  
It factorizes the raw inputs into the object-centric representations, converts them into probabilistic ground atoms, and finally performs differentiable forward-chaining inference using weighted rules.
 NSFR works as follows, cf. Fig.~\ref{fig:workflow}:

\noindent
{\bf Step 1:} Let $\mathbf{X} \in \mathbb{R}^{B \times N}$ be a batch of input images. Perception function $f_\text{per}^{\bm{\Phi}}: \mathbb{R}^{B \times N} \rightarrow \mathbb{R}^{B \times E \times D}$ factorizes input ${\bf X}$ into a set of object-centric representations $\mathbf{Z} \in \mathbb{R}^{B \times E \times D}$, where  $E \in \mathbb{N}$ is the number of objects, and $D \in \mathbb{N}$ is the dimension of the object-centric vector.
\\
{\bf Step 2:} Let $\mathcal{G}$ be a set of ground atoms.
Convert function $f_\text{con}^{\bm{\Theta}}: \mathbb{R}^{B \times E \times D} \times \mathcal{G} \rightarrow \mathbb{R}^{B \times G}$ generates a probabilistic vector representation of facts, where $G = |\mathcal{G}|$.
\\
{\bf Step 3:} Infer function $f_\text{inf}^{\bm{W}}: \mathbb{R}^{B \times G} \rightarrow \mathbb{R}^{B \times G}$ computes forward-chaining inference using weighted clauses $\mathcal{C}$.\\
{\bf Step 4:} Predict function $f_\text{pre}: \mathbb{R}^{B \times G} \rightarrow \mathbb{R}^B$ computes the probability of target facts. The probability of the labels $\mathbf{y}$ of the batch of input ${\bf X}$ is computed as:
\begin{align*}
	p(\mathbf{y}|\mathbf{X}) = f_{pre} ( f_{inf} ( f_{con} ( f_{per} (\mathbf{X};  \bm{\Phi}), \mathcal{G}; \bm{\Theta}); \mathcal{C}, \mathbf{W})),
\end{align*}
where $\bm{\Phi}, \bm{\Theta}$, and $\mathbf{W}$ are learnable parameters.
Essentially, NSFR computes the logical entailment for given weighted clauses and probabilistic atoms in a differentiable way. 
As increasing number of attributes lead to vanishing probabilities, we normalize the final result taking the $n$-th square root:  $\Tilde{p}(\mathbf{y}|\mathbf{X}) = \sqrt[n]{p(\mathbf{y}|\mathbf{X})} \approx \sqrt[n]{\Pi_i^n(p_i)}$, with $p_i$ being the $i$-th found attribute.

Next we will provide empirical evidence demonstrating that LogicRank  enables text-to-image models like DALL-E  to reliably generate images from precise statements.

\section{Empirical Evaluation}
%\section{Logical DrawBench}
\label{sec:comparison}

We now compare the workflows of CLIP with its scores as a reranker (Fig.~\ref{fig:dalle_workflow}), and LogicRank with its probability estimates (Fig.~\ref{fig:workflow}) respectively.
Inspired by the \textit{draw bench} benchmark~\cite{imagen}, we consider three groups of tasks  which require logical reasoning, namely \textit{coloring}, \textit{positioning} and \textit{counting}.
We start by comparing the reranking modules CLIP and LogicRank isolated from the generative model. To this end, we investigate the task of counting objects in given, i.e., not generated, real images.
Next, we evaluate the complete pipeline: generating images with DALL-E mini and comparing the top reranked samples utilizing CLIP vs. LogicRank.

% More precisely we evaluate:
% \begin{itemize}
%     \item counting of the number of dogs in images of the \mbox{COCON} dataset
%     \item generation of images with DALL-E, containing spheres and cubes, of certain colors, numbers, with relative positions and reranking those
% \end{itemize}

\subsection{Object Counting}
\begin{figure}[t]
\centering
\begin{subfigure}[b]{\linewidth}
\includegraphics[width=.49\linewidth]{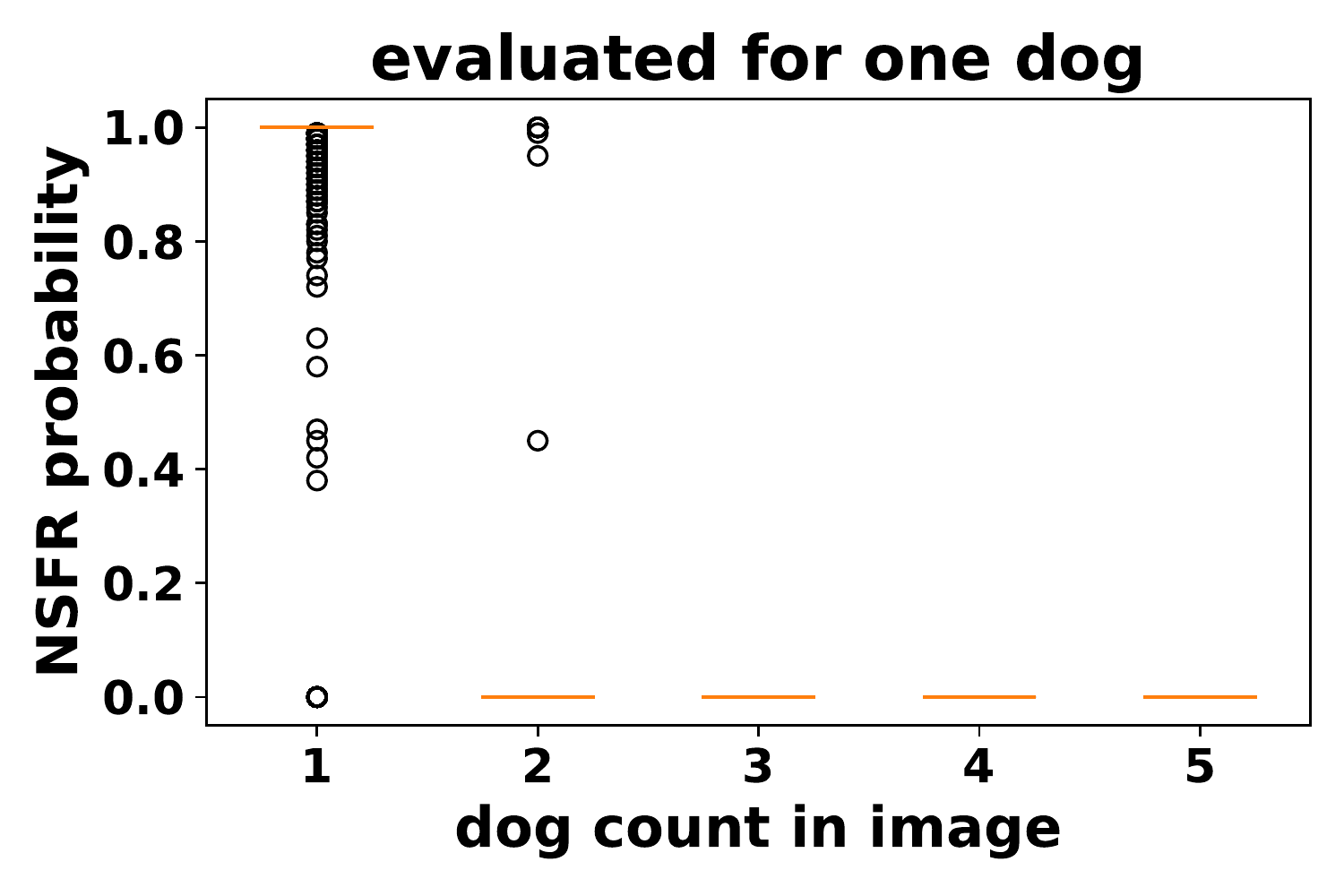}
\includegraphics[width=.49\linewidth]{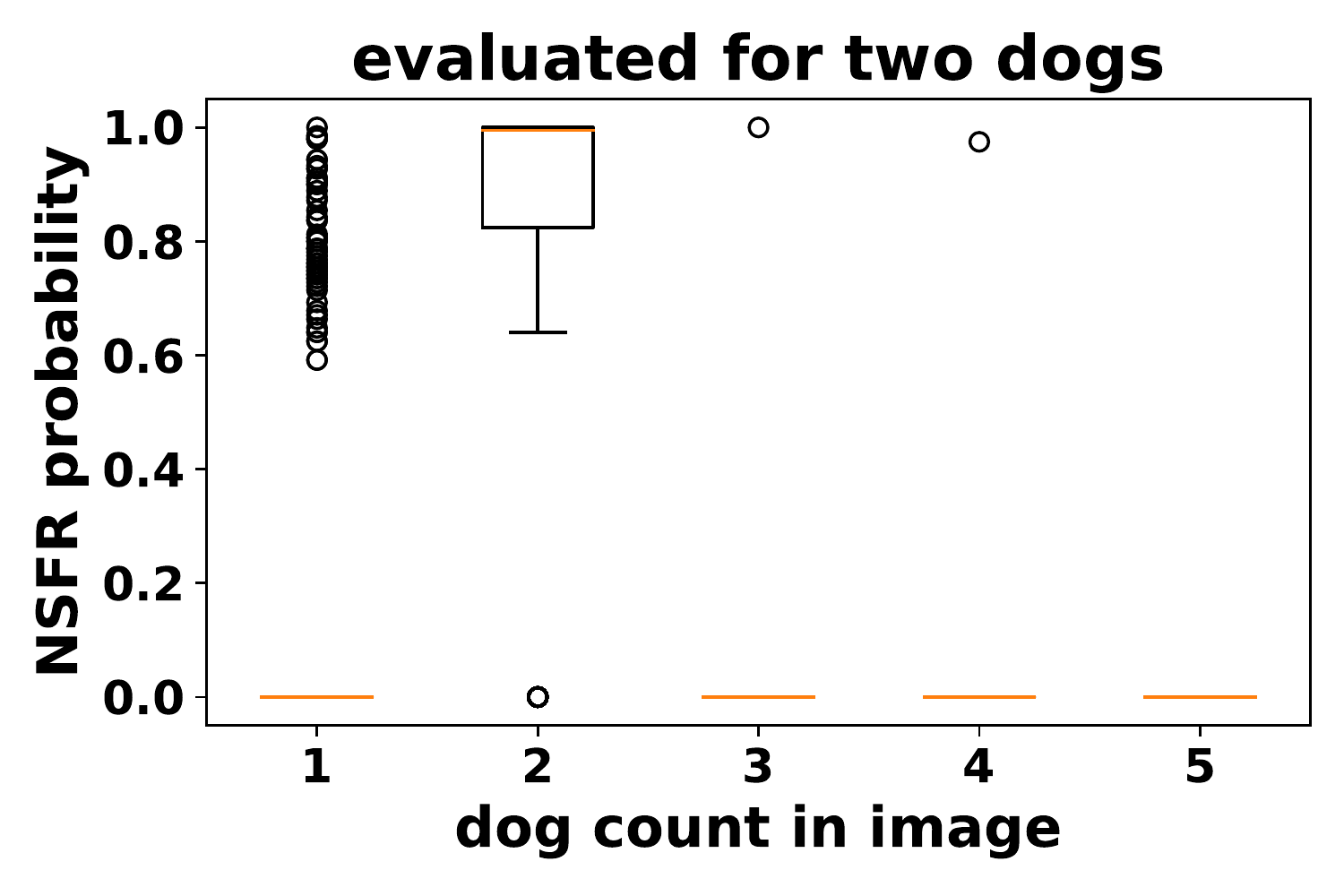}
\end{subfigure}
\begin{subfigure}[b]{\linewidth}
\includegraphics[width=.49\linewidth]{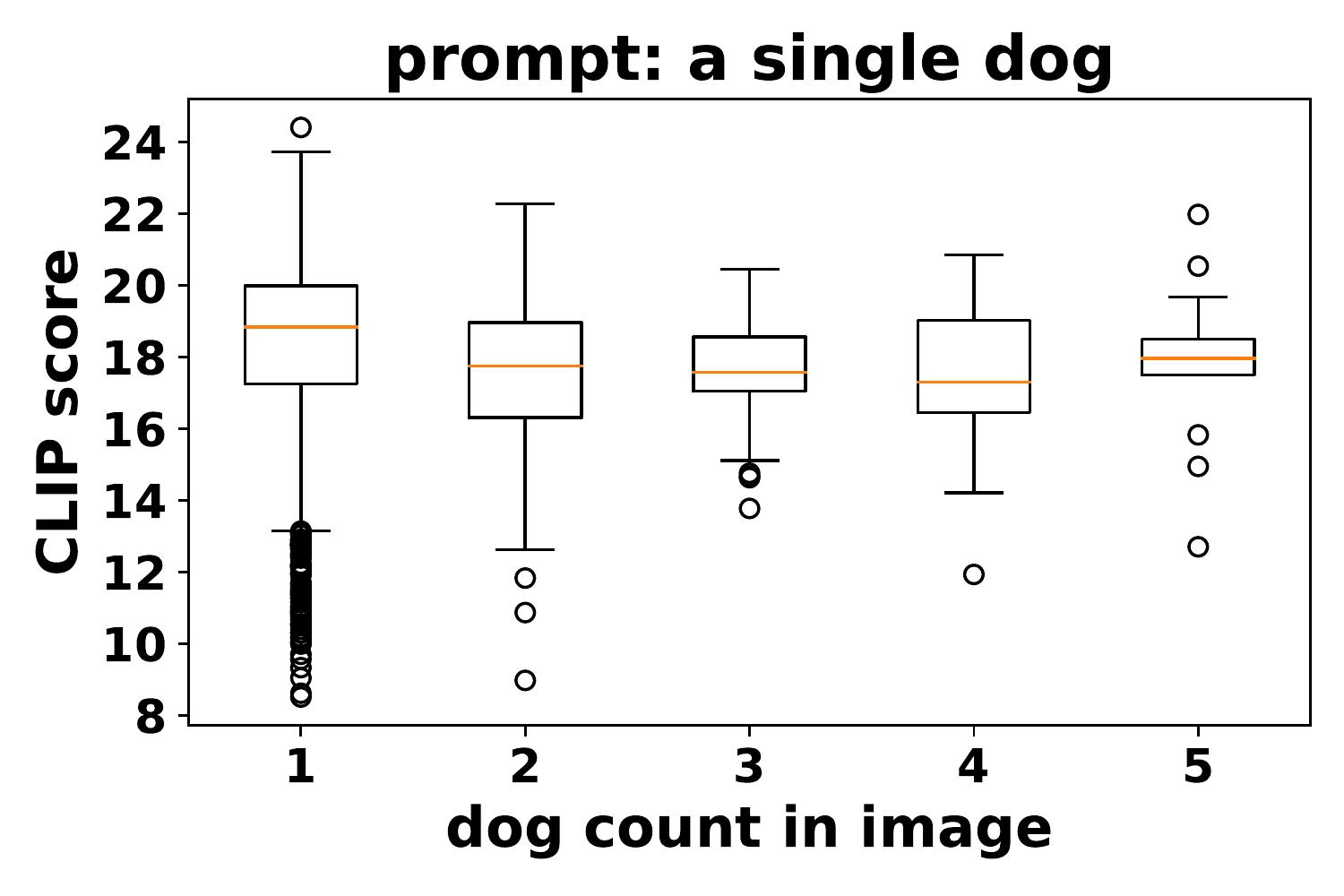}
\includegraphics[width=.49\linewidth]{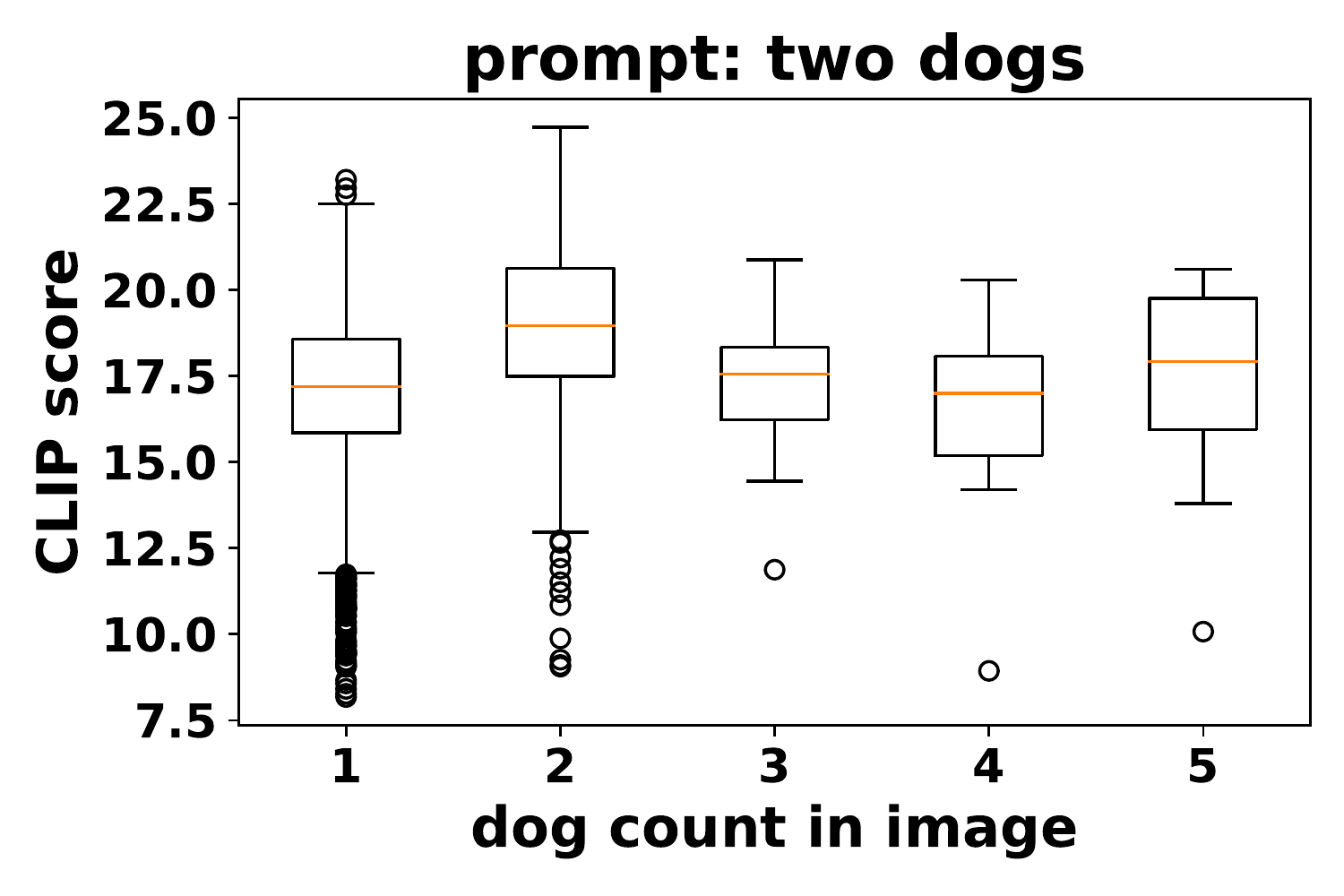}
\end{subfigure}
\caption{Boxplots evaluating NSFR-probabilities and CLIP-scores (y-axis) for image sets containing different numbers of dogs of the COCON dataset (x-axis). 
For CLIP we show results for the best text prompts separating the categories one and two dogs. While the score tendencies in principle match with the correct class, it can be observed that other groups produce significantly overlapping scores, rendering CLIP by itself not sufficient as a precise reranker.
For NSFR, it can be seen that only few outliers from neighboring groups produce high probabilities.}
\label{fig:inconsistent2} 
\label{fig:nsfr_dogs}
\end{figure}

We start our evaluation with a simple counting benchmark comparing the CLIP ranking and the LogicRank ranking. 
For this evaluation we first group the images of the \mbox{MS COCON} dataset~\cite{coco} by the number of dogs contained in their annotations. 
In the bottom row of Fig.~\ref{fig:inconsistent2} we show the produced CLIP scores (y-axis) on such groups (x-axis). The used text-prompt for separation with CLIP\footnote{We only show the best prompt out of several hand-crafted choices.} of the class in choice is printed in the title. While it can be observed that the boxplot of the matching class is in principle better represented, neighboring classes produce significant overlap, therefore rendering the standard CLIP model as a bad candidate for a reranking  of such counting tasks. 

The top row in comparison shows the same experiment evaluated on probabilities of LogicRank. 
As an object perception module we use the pre-trained DETR. As logical rules, we use $kp_i(im) \textrm{:-} \bigwedge_{j=0..i} contain(im, \textrm{dog}) \wedge \neg \bigwedge_{j=0..(i+1)} contain(im, \textrm{dog})$ for separating the $i$-th group.
The proposed approach is clearly able to distinguish the categories except for some outliers.\footnote{Outliers, of which some are due to false annotations.}

\subsection{Demanding Logical Reasoning}
\begin{figure}[th!]
\centering
\begin{tabular}[t]{p{.97\linewidth}}
Task1: ``one red cube and one blue sphere right of it''\\
%\vspace*
\begin{subfigure}[b]{\linewidth}
\includegraphics[width=\linewidth]{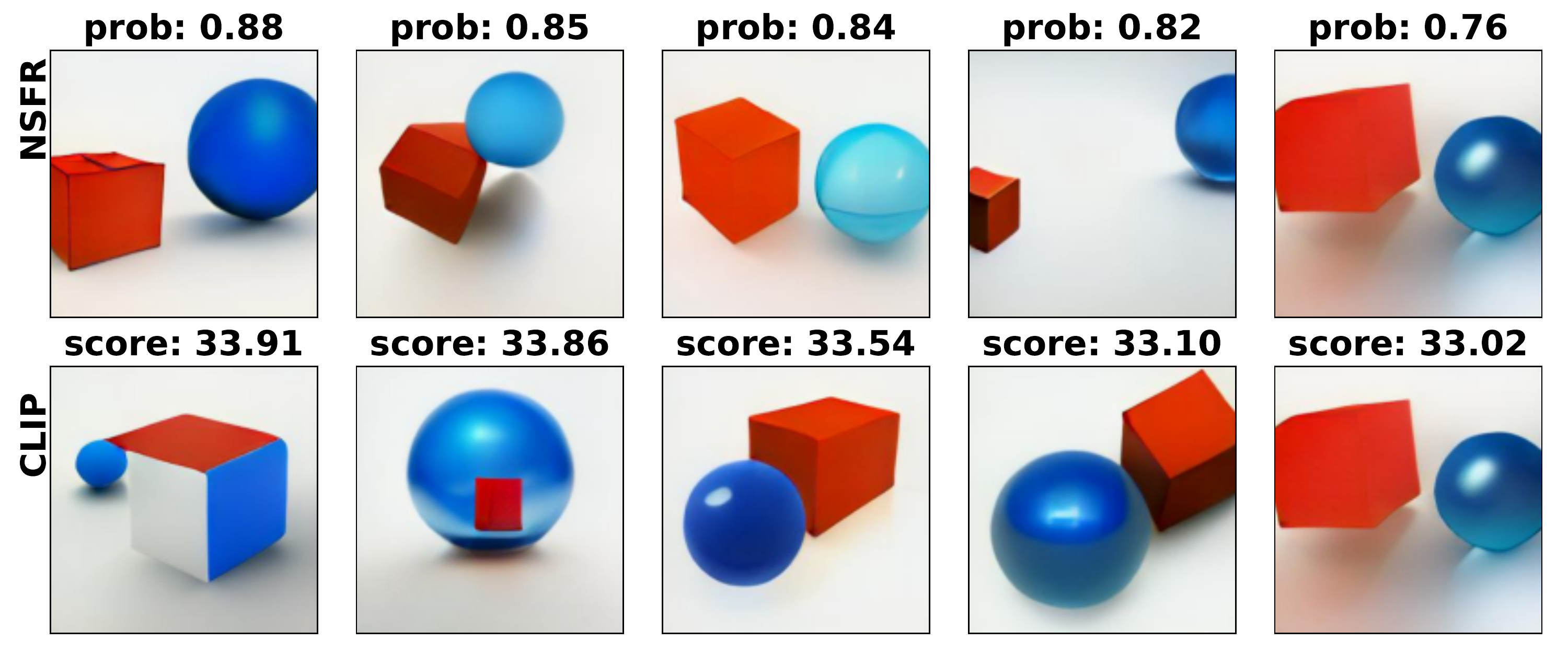}
\end{subfigure}\\ %clip_1
\vspace{1pt}Task2: ``one blue cube and one red sphere right of it''\\
\vspace*{-8pt}
\begin{subfigure}[b]{\linewidth}
\includegraphics[width=\linewidth]{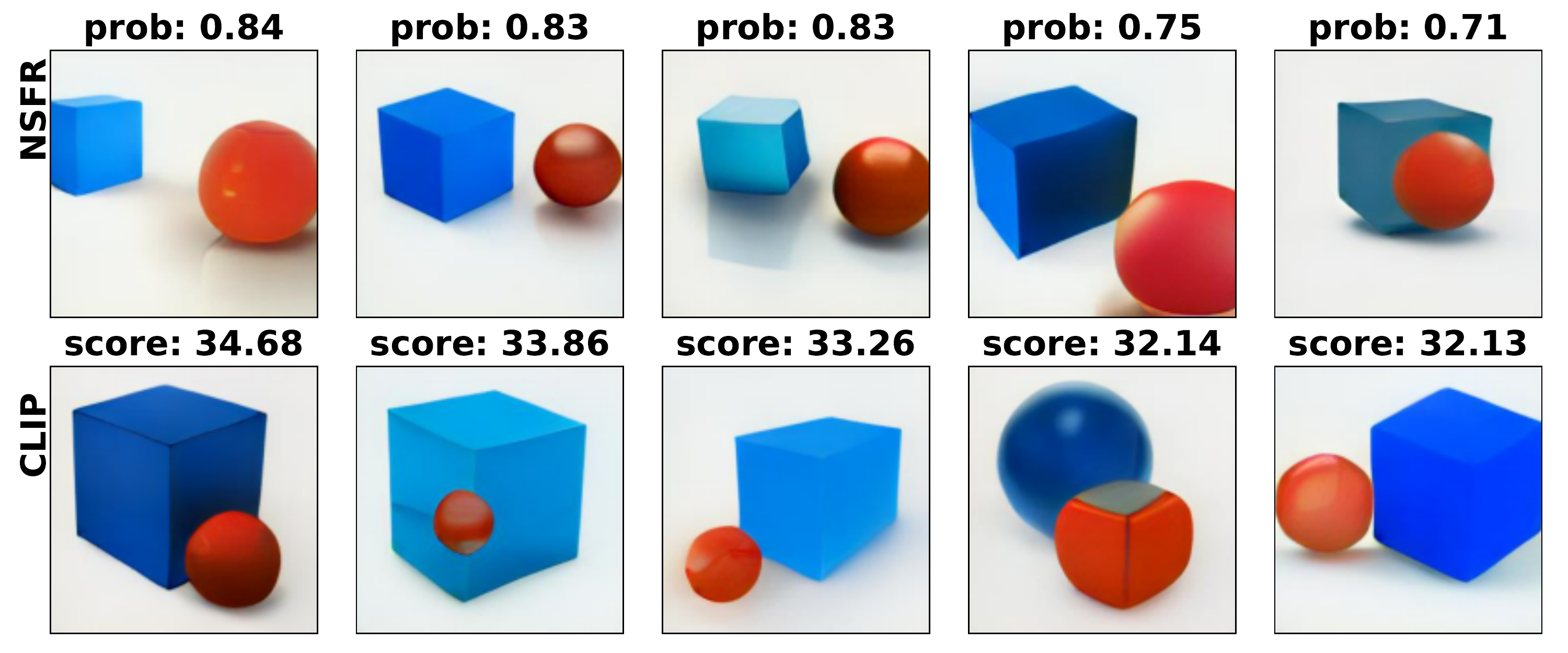}
\end{subfigure}\\   %clip_2
\vspace{1pt}Task3: ``two blue spheres and one red cube''\\
\vspace*{-8pt}
\begin{subfigure}[b]{\linewidth}
\includegraphics[width=\linewidth]{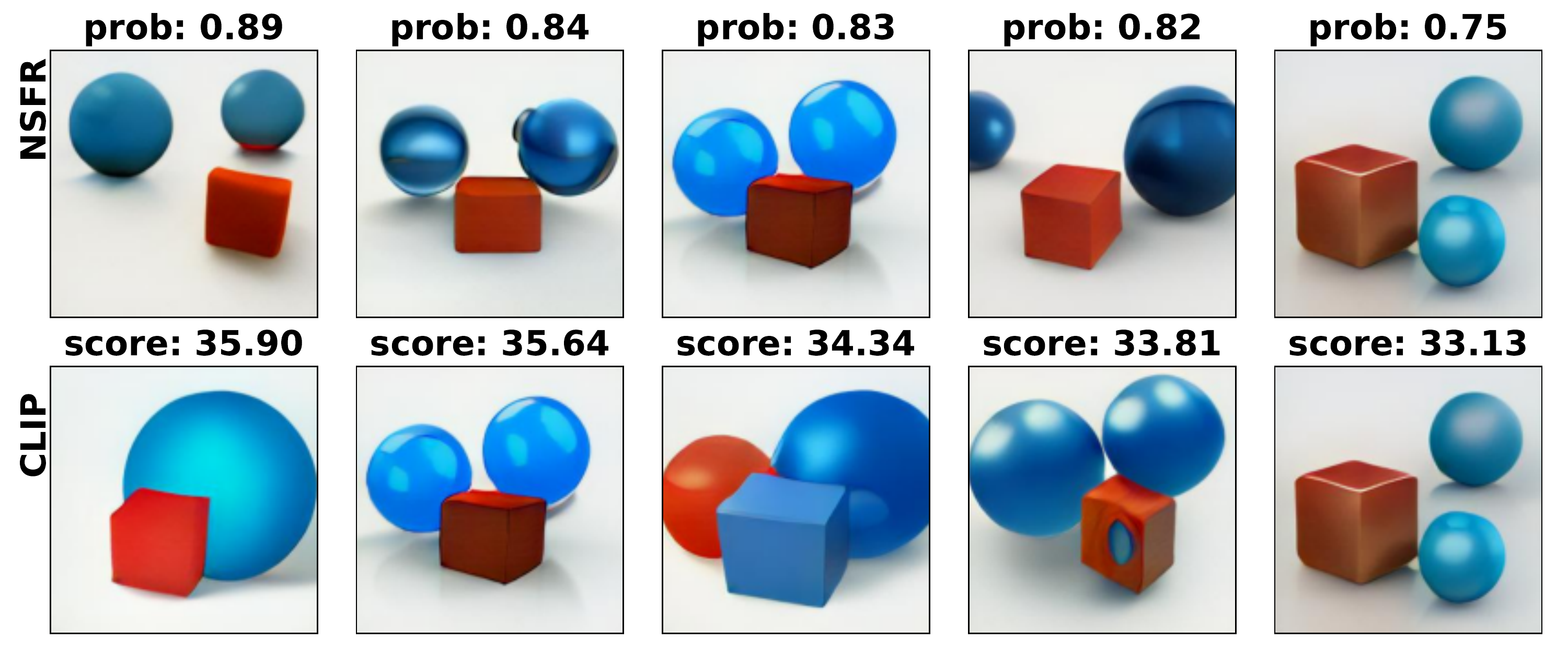}
\end{subfigure}\\  %task3
\vspace{1pt}Task4$^*$: ``at least two blue spheres and a red cube, the biggest sphere below a red cube''\\
\vspace*{-8pt}
\begin{subfigure}[b]{\linewidth}
\includegraphics[width=\linewidth]{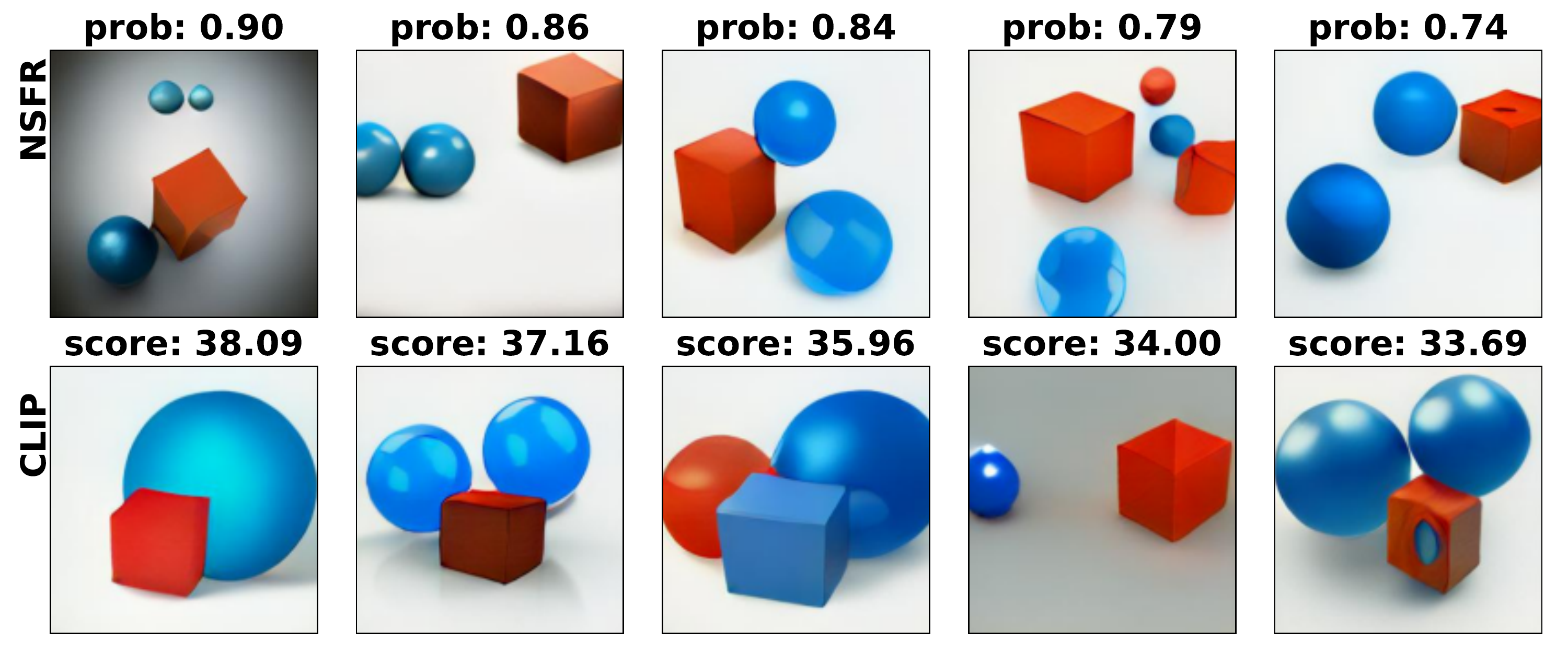}
\end{subfigure}%compare_exp3
\end{tabular}
\captionof{figure}{Top-5 (from left to right) results for several statements, each evaluated with NSFR (top-row) and CLIP (bottom-row) respectively, as described with the workflows Fig.~\ref{fig:dalle_workflow} and \ref{fig:workflow} above. For each statement, 200 samples were generated, except for Task4, as described in text. One can observe that NSFR consistently matches the task description through all tasks, while CLIP confuses count, color, shape and positioning of objects.
}
\label{fig:compare0}
\label{fig:compare1}
\label{fig:compare2}
\label{fig:compare3}
\end{figure}

Next, we move to more complex queries requiring logical reasoning in the text-to-image workflow and analyse the reranking within the generation process.
For the entire pipelines of Fig.~\ref{fig:dalle_workflow} and \ref{fig:workflow} we generate images with DALL-E mini\footnote{\nopagebreak {We stayed with $\text{tmp}=0$ and $\text{cond\_scale}=3.0$ as proposed by the authors, as we found it producing best results.}} and evaluate them with CLIP and LogicRank on the same statements respectively.
To compare the ranking methods, we show the top-5 candidates for each method  in Fig.~\ref{fig:compare0} for 4 different statements. For each statement we generate 200 samples to also obtain enough qualitative images fitting the description. The candidates are then reranked, in the top-rows with our proposed method on the logical rule, and in the bottom-row with CLIP on the text-prompt.

For Task1-3, it can be observed that our LogicRank approach  stays throughout consistent with the task description, while CLIP mixes up color to shape as in Task1 example 1 (from left to right) and 2, positioning as in Task2 example 3, or even the right amount of objects as in Task3 example 1.
For the statement of Task4, DALL-E did not create any useful footage. We therefore generated 1500 samples with the statement of Task3, and just reranked them with CLIP and LogicRank respectively. Note that we did not explicitly ask DALL-E to generate images with more than 3 objects, even though we find some with NSFR.
Comparing Task3 and 4, it can moreover be observed that CLIP favors the same samples over different statements disregarding the precise text-prompt. Example 1 even remains at the top position\footnote{Even for Task1 and Task2, example 1 of Task3  scores the highest with $39.02$ and $38.41$ respectively- leaving  some distance to the other, more precise samples.}. For LogicRank, all 5 results match the description. 
Overall our method proofs more robust for selecting top samples, even for fairly complex rules.

The CLIP score clearly is not usable as a proper class separator. Throughout the top-5 candidates of all tasks, it remains within a small range giving no hint about the actual missing details, even though the quality differs a lot.  
As observed with Fig.~\ref{fig:finetuned} above, with the proposed LogicRank approach we can unravel the specific attributes and obtain their estimated probabilities. In the scenario of diffusion models, this precise feedback could potentially be used as guidance for specific re-sampling of object-attributes and locations. 

\section{Conclusion}
%\section{Towards an advanced\\ Logical Reasoning Benchmark.}
\label{sec:conclusion}
In this work we demonstrated how language-image multi-models, in particular DALL-E and CLIP, suffer with accurate statements in generation processes as well as for similarity matching.
For more accurate text-image similarity matching, we proposed LogicRank. It  utilizes  an object detector combined with Neuro-Symbolic Forward Reasoning to derive matching probability estimates. The latter approach is clearly able to fulfill the ``counting'', ``positional'' and ``colors'' categories of the DrawBench prompts~\cite{imagen}. 
More complex rules could consistently be ranked, even for DALL-E generated images unseen before when fine-tuning the DETR model.

In ongoing research we first want build a benchmark dataset to evaluate DrawBench like precision capabilities utilizing LogicRank.
We then plan to integrate the reasoner and its probability estimate as additional quality feedback during the diffusion process of DALL-E 2~\cite{dalle2}. We believe it can be used to efficiently guide the model during training as well as inference towards a more precise and expected results. This method should prove more efficient, as it can directly refer to the image and attribute in question. 

\subsection*{Acknowledgments}
 This research has benefited from the Hessian Ministry of Higher Education, Research, Science and the Arts (HMWK) cluster project ``The Third Wave of AI''.

\newpage

\bibliography{references}

\end{document}